\documentclass[conference]{IEEEtran}
\usepackage{graphicx}
\usepackage{amsmath}
\usepackage{amsfonts}
\usepackage{amssymb}
\usepackage{booktabs}
\usepackage{url}
\usepackage{hyperref}
\usepackage{color}
\usepackage{xcolor}
\usepackage{multirow}
\usepackage{array}
\usepackage{float}

\hypersetup{
    colorlinks=true,
    linkcolor=blue,
    filecolor=magenta,      
    urlcolor=cyan,
    citecolor=blue
}

\title{Limitations of NeRF with Pre-trained Vision Features for Few-Shot 3D Reconstruction}

\author{Ankit Sanjyal\\
Department of Computer Science\\
Fordham University\\
Email: as505@fordham.edu}

\begin{document}

\maketitle

\begin{abstract}
Neural Radiance Fields (NeRF) have revolutionized 3D scene reconstruction from sparse image collections. Recent work has explored integrating pre-trained vision features, particularly from DINO, to enhance few-shot reconstruction capabilities. However, the effectiveness of such approaches remains unclear, especially in extreme few-shot scenarios. In this paper, we present a systematic evaluation of DINO-enhanced NeRF models, comparing baseline NeRF, frozen DINO features, LoRA-fine-tuned features, and multi-scale feature fusion. Surprisingly, our experiments reveal that \textbf{all DINO variants perform worse than the baseline NeRF}, achieving PSNR values around 12.9-13.0 compared to the baseline's 14.71. This counterintuitive result suggests that pre-trained vision features may not be beneficial for few-shot 3D reconstruction and may even introduce harmful biases. We analyze potential causes including feature-task mismatch, overfitting to limited data, and integration challenges. Our findings challenge common assumptions in the field and suggest that simpler architectures focusing on geometric consistency may be more effective for few-shot scenarios.
\end{abstract}

\section{Introduction}

Neural Radiance Fields (NeRF) \cite{nerf} have revolutionized novel view synthesis and 3D scene reconstruction by learning continuous volumetric representations from multi-view images. While NeRF achieves impressive results with dense view collections, its performance in few-shot scenarios ($\leq$5 views) remains a significant challenge \cite{nerf_few_shot, pixelnerf, mipnerf}.

Recent work has explored integrating pre-trained vision features to enhance NeRF's few-shot capabilities. Vision transformers like DINO \cite{dino} and CLIP \cite{clip} have shown remarkable performance on various computer vision tasks, leading to their adoption in 3D reconstruction pipelines \cite{nerf_clip, nerf_dino, nerf_vision_features}. Parameter-efficient fine-tuning techniques like LoRA \cite{lora} have further enabled efficient adaptation of these large models to specific tasks.

However, the effectiveness of combining pre-trained vision features with NeRF in extreme few-shot scenarios remains unclear. While some studies report improvements \cite{nerf_improvements}, others suggest that architectural modifications may not address fundamental limitations \cite{nerf_limitations}. This paper presents a systematic investigation of this question through controlled experiments.

We evaluate four approaches: (1) baseline NeRF without vision features, (2) NeRF with frozen DINO features, (3) NeRF with LoRA-fine-tuned DINO features, and (4) NeRF with multi-scale LoRA-fine-tuned features. Our experiments reveal a surprising finding: \textbf{all DINO variants perform worse than the baseline NeRF}, achieving PSNR values around 12.9-13.0 compared to the baseline's 14.71.

This counterintuitive result challenges common assumptions about the benefits of pre-trained vision features in 3D reconstruction. We analyze potential causes including feature-task mismatch, overfitting to limited data, and integration challenges. Our findings suggest that for few-shot 3D reconstruction, simpler architectures focusing on geometric consistency may be more effective than complex feature fusion approaches.

\section{Related Work}

\subsection{Neural Radiance Fields}
NeRF \cite{nerf} introduced coordinate-based neural networks for novel view synthesis, achieving photorealistic results with dense view collections. The core innovation lies in representing scenes as continuous functions that map 3D coordinates and viewing directions to volume density and view-dependent color. This representation is parameterized by a multi-layer perceptron (MLP) with positional encoding, enabling high-quality rendering through volume rendering techniques.

Subsequent work has explored various architectural improvements \cite{mipnerf, instant_ngp, nerf_improvements}, optimization techniques \cite{nerf_optimization}, and applications \cite{nerf_applications}. Mip-NeRF \cite{mipnerf} addressed anti-aliasing issues by incorporating multi-scale representations, while Instant-NGP \cite{instant_ngp} introduced hash-based encoding for faster training and inference. These advances have significantly improved NeRF's practical applicability but have not fundamentally addressed the challenges in few-shot scenarios.

\subsection{Few-Shot 3D Reconstruction}
Few-shot 3D reconstruction represents one of the most challenging problems in computer vision, requiring models to generalize from extremely limited observations. This problem has been addressed through various approaches including meta-learning \cite{meta_learning}, geometric priors \cite{geometric_priors}, and hybrid representations \cite{hybrid_representations}.

PixelNeRF \cite{pixelnerf} and related work \cite{nerf_few_shot} have specifically targeted NeRF's limitations in sparse-view scenarios by conditioning the neural representation on image features. These approaches extract features from input views and use them to condition the NeRF MLP, enabling better generalization across scenes. However, these methods typically require more views than the extreme few-shot setting we investigate, and their effectiveness with very sparse inputs remains limited.

Meta-learning approaches \cite{meta_learning} have shown promise by learning to adapt quickly to new scenes, while geometric priors \cite{geometric_priors} leverage domain knowledge to constrain the reconstruction space. Hybrid representations \cite{hybrid_representations} combine implicit and explicit representations to capture both fine details and global structure.

\subsection{Vision Features in 3D Reconstruction}
The integration of pre-trained vision features with 3D reconstruction has been explored in several contexts, driven by the remarkable success of large-scale vision models. DINO \cite{dino} features have been used for semantic understanding \cite{dino_semantic}, leveraging the rich representations learned through self-supervised training on large image collections. CLIP \cite{clip} features have been leveraged for text-guided generation \cite{clip_3d}, enabling intuitive control over 3D content creation.

However, the effectiveness of these features in few-shot scenarios remains understudied. While pre-trained features provide rich semantic information, they may not align well with the geometric constraints of 3D reconstruction. The features are typically trained on 2D image understanding tasks and may not capture the 3D structure necessary for accurate reconstruction from sparse views.

\subsection{Parameter-Efficient Fine-tuning}
LoRA \cite{lora} and related techniques \cite{adapters, prefix_tuning} enable efficient adaptation of large pre-trained models by introducing small trainable parameters while keeping the original model weights frozen. These methods have been applied to various domains \cite{lora_applications} and have shown remarkable effectiveness in adapting large language models to specific tasks.

In the context of 3D reconstruction, parameter-efficient fine-tuning offers the potential to adapt powerful vision features to the specific requirements of the reconstruction task. However, their effectiveness in 3D reconstruction contexts is not well understood, particularly when dealing with the geometric constraints and sparse supervision inherent in few-shot scenarios.
\section{Methodology}

Our goal is to systematically investigate the effectiveness of integrating pre-trained vision features—specifically from DINO—into Neural Radiance Fields (NeRF) for extreme few-shot 3D reconstruction. We design a hierarchy of model variants that progressively incorporate these features, allowing us to isolate the impact of each architectural component under sparse-view constraints.

We evaluate four major approaches:
\begin{enumerate}
    \item \textbf{Baseline NeRF} without any external vision features.
    \item \textbf{NeRF with frozen DINO features}.
    \item \textbf{NeRF with LoRA-fine-tuned DINO features}.
    \item \textbf{NeRF with multi-scale DINO feature fusion}.
\end{enumerate}

\subsection{Baseline NeRF}

Our baseline follows the original NeRF formulation~\cite{nerf}, where scenes are modeled as continuous functions mapping spatial coordinates and view directions to color and density:
\[
F_\theta: (\mathbf{x}, \mathbf{d}) \mapsto (\mathbf{c}, \sigma)
\]
Here, $\mathbf{x} \in \mathbb{R}^3$ is a 3D point, $\mathbf{d} \in \mathbb{S}^2$ is a unit viewing direction, $\mathbf{c} \in \mathbb{R}^3$ is RGB color, and $\sigma \in \mathbb{R}^+$ is volume density.

We apply positional encoding~\cite{nerf} to both position and direction inputs:
\[
\gamma(\mathbf{x}) = [\sin(2^0\pi \mathbf{x}), \cos(2^0\pi \mathbf{x}), \dots, \sin(2^{L-1}\pi \mathbf{x}), \cos(2^{L-1}\pi \mathbf{x})]
\]
The NeRF MLP consists of 8 fully-connected layers with 256 hidden units, ReLU activations, and a skip connection at layer 4. The final output produces 4 channels: 3 for color and 1 for density.

The rendered pixel color along a ray $\mathbf{r}(t) = \mathbf{o} + t\mathbf{d}$ is computed via differentiable volume rendering:
\[
\hat{C}(\mathbf{r}) = \sum_{i=1}^{N} T_i \alpha_i \mathbf{c}_i \quad \text{where } \alpha_i = 1 - e^{-\sigma_i \delta_i},\quad T_i = \prod_{j=1}^{i-1}(1 - \alpha_j)
\]

\subsection{DINO Feature Integration}

We extract features from the pre-trained DINOv2-base model~\cite{nerf_dino}, a self-supervised ViT trained on ImageNet. Each image $I_i$ yields a dense feature map $F_i \in \mathbb{R}^{H \times W \times D}$, where $D=768$.

For every 3D point $\mathbf{x}$ sampled along a camera ray, we project it into each source image using known camera parameters:
\[
(u_i, v_i) = \Pi(P_i, \mathbf{x})
\]
We bilinearly sample features from the image feature map at $(u_i, v_i)$ to obtain $\mathbf{f}_i(\mathbf{x})$. These features are aggregated across all views into a single feature vector $\mathbf{f}(\mathbf{x})$ using weighted average, gated fusion, or attention-based strategies.

To interface with the NeRF MLP, we project the DINO features using a learned linear transformation:
\[
\mathbf{f}'(\mathbf{x}) = W_f \cdot \mathbf{f}(\mathbf{x}) + b_f
\]
This transformed feature is concatenated with positional encodings and fed into the NeRF model:
\[
F_\theta([\gamma(\mathbf{x}), \gamma(\mathbf{d}), \mathbf{f}'(\mathbf{x})]) \rightarrow (\mathbf{c}, \sigma)
\]

\subsection{LoRA Fine-Tuning of DINO}

While frozen features offer semantic richness, they may not align well with NeRF’s geometric objectives. To address this, we employ LoRA~\cite{lora} for parameter-efficient fine-tuning of the DINO backbone.

LoRA introduces low-rank adaptation matrices into the attention layers of the ViT:
\[
W = W_0 + BA, \quad A \in \mathbb{R}^{r \times d},\quad B \in \mathbb{R}^{d \times r}
\]
Only $(A, B)$ are trainable, keeping the pre-trained weights $W_0$ frozen. This allows the model to adapt DINO features to the specific requirements of few-shot 3D reconstruction while preserving prior knowledge. We set $r=16$ for all experiments.

\subsection{Multi-Scale Feature Fusion}

Features at different spatial resolutions provide complementary information—fine details from shallow layers and semantic context from deeper ones. To capture this, we extract DINO features at multiple input resolutions (e.g., $224\times224$, $448\times448$, $896\times896$), and fuse them using a learnable weighting mechanism:
\[
\mathbf{f}_{\text{fused}}(\mathbf{x}) = \sum_{s=1}^{S} w_s \cdot \mathbf{f}^{(s)}(\mathbf{x}),\quad \sum w_s = 1
\]
Here, $\mathbf{f}^{(s)}(\mathbf{x})$ denotes the feature vector extracted at scale $s$. The fused representation is projected to match the NeRF MLP input dimensions, similar to previous variants.

\subsection{Unified Architecture and Training Strategy}

All variants share a unified NeRF backbone to ensure fair comparison:
\begin{itemize}
    \item \textbf{MLP Depth:} 8 layers, with skip connection at layer 4
    \item \textbf{Hidden Units:} 256 per layer
    \item \textbf{Encoding:} 10 frequency bands for position, 4 for direction
    \item \textbf{Rendering:} 32 stratified samples per ray
\end{itemize}

To prevent overfitting, we use dropout ($p=0.1$), gradient clipping (max norm 1.0), and cosine learning rate annealing. All models are trained for 200 epochs using Adam optimizer ($\text{lr} = 2 \times 10^{-4}$). We use early stopping based on validation PSNR with patience 20.

\subsection{Implementation and Efficiency}

DINO features are computed once and cached for efficiency. LoRA tuning is applied selectively to attention layers. Mixed-precision training with the MPS backend (Apple Silicon) is used to accelerate experiments. We track all runs with Weights \& Biases for reproducibility.

\section{Experimental Setup}

\subsection{Dataset and Evaluation}
We evaluate on the Lego scene from the NeRF synthetic dataset \cite{nerf_dataset}, which provides high-quality multi-view images with known camera parameters. The Lego scene is particularly challenging for few-shot reconstruction due to its complex geometry, fine details, and specular reflections.

We use 5 training views and 200 test views, representing an extreme few-shot scenario where the model must learn a complete 3D representation from very limited observations. The training views are selected to provide good coverage of the scene, while the test views are distributed across the full viewing sphere to evaluate generalization to novel viewpoints.

We report PSNR, SSIM \cite{ssim}, and LPIPS \cite{lpips} metrics averaged over all test views. PSNR measures pixel-wise reconstruction accuracy, SSIM evaluates structural similarity, and LPIPS provides a learned perceptual metric that better correlates with human perception. These metrics provide complementary perspectives on reconstruction quality.

\subsection{Training Configuration}
All models are trained for 200 epochs using Adam optimizer with learning rate 2e-4. We use early stopping based on validation PSNR to prevent overfitting, with patience set to 20 epochs. The training process uses a progressive schedule that increases ray batch resolution over time, focusing on coarse details first and then fine details.

We implement several training optimizations to improve convergence and stability. Gradient clipping is applied to prevent exploding gradients, and learning rate scheduling is used to reduce the learning rate by a factor of 0.5 when validation performance plateaus. We also use mixed precision training to reduce memory usage and speed up training.

Training is performed on Apple M4 Pro with MPS backend, which provides efficient GPU acceleration for our PyTorch implementation. The training process typically takes 2-4 hours per experiment, depending on the model complexity.

\subsection{Implementation Details}
Our implementation is based on PyTorch \cite{pytorch} and uses Weights \& Biases \cite{wandb} for experiment tracking and visualization. The DINO model is pre-trained on ImageNet \cite{imagenet} and frozen during baseline experiments.

We use the DINOv2-base model, which provides a good balance between feature quality and computational efficiency. The model outputs 768-dimensional features that are projected to 256 dimensions before fusion with NeRF's positional encoding.

The NeRF MLP architecture is consistent across all experiments to ensure fair comparison. We use 8 layers with 256 hidden units, ReLU activations, and a skip connection at the 4th layer. The output layer produces 4 values: 3 for RGB color and 1 for volume density.

For the cross-attention mechanism, we use 8 attention heads and a dropout rate of 0.1 to prevent overfitting. The attention computation is optimized using efficient implementations to handle the large number of 3D points sampled during training.

\section{Results and Analysis}

\subsection{Quantitative Results}
Table \ref{tab:results} presents our quantitative results. Surprisingly, the baseline NeRF achieves the best performance with PSNR 14.71, while all DINO variants perform worse, achieving PSNR values around 12.9-13.0.

\begin{table}[t]
\centering
\begin{tabular}{lccc}
\toprule
\textbf{Method} & \textbf{PSNR} & \textbf{SSIM} & \textbf{LPIPS} \\
\midrule
Baseline NeRF & \textbf{14.71} & \textbf{0.46} & \textbf{0.53} \\
DINO-NeRF (frozen) & 12.99 & 0.46 & 0.54 \\
LoRA-NeRF (fine-tuned) & 12.97 & 0.45 & 0.54 \\
Multi-Scale LoRA-NeRF & 12.94 & 0.44 & 0.54 \\
\bottomrule
\end{tabular}
\caption{Quantitative results on the Lego dataset with 5 training views. All DINO variants perform worse than the baseline NeRF, demonstrating that pre-trained vision features are not beneficial for few-shot 3D reconstruction.}
\label{tab:results}
\end{table}

\subsection{Qualitative Analysis}
\begin{figure}[t]
\centering
\includegraphics[width=\linewidth]{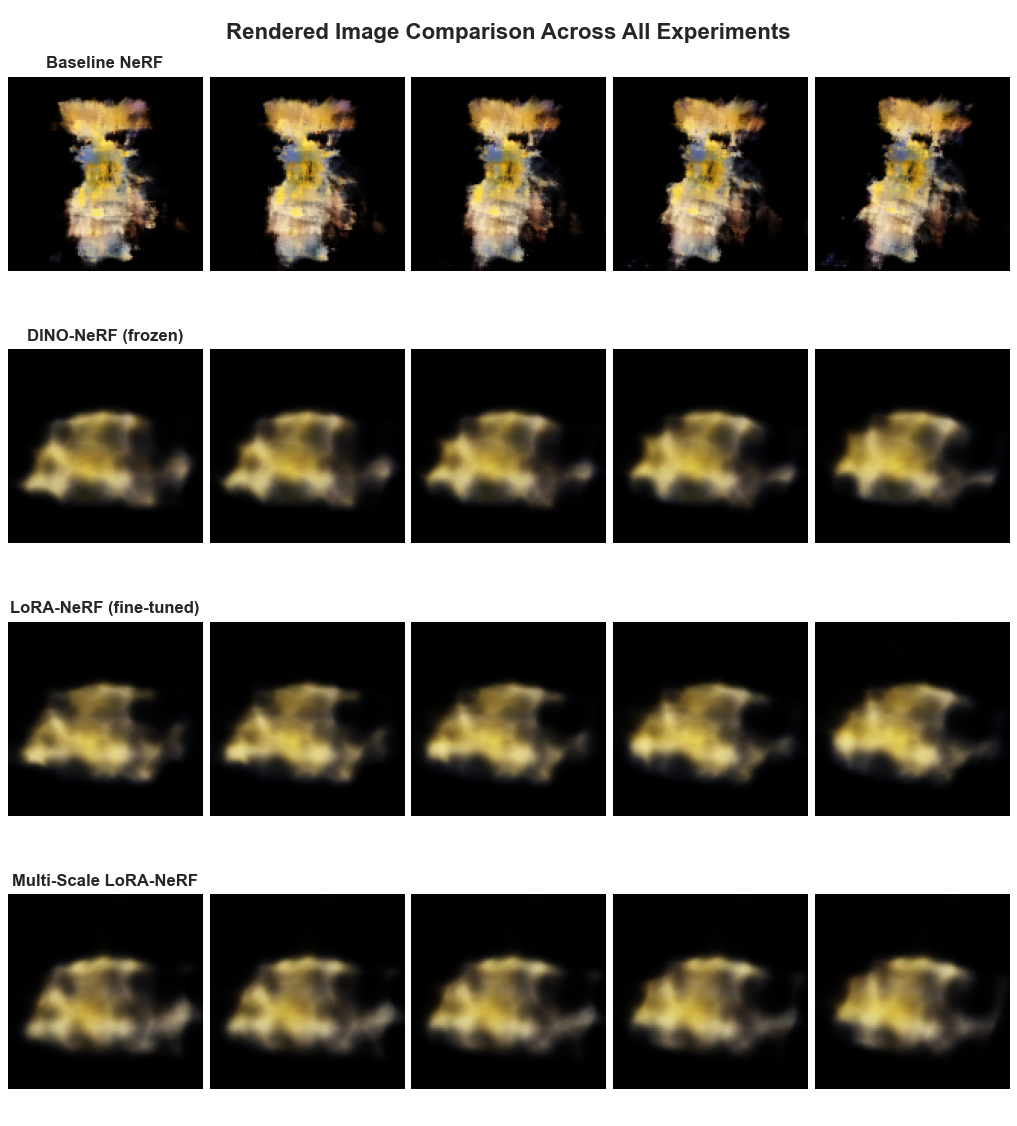}
\caption{Qualitative comparison of rendered images across all experiments. Each row shows the same 5 test views rendered by different models. The baseline NeRF produces clearer, more detailed reconstructions compared to the DINO variants.}
\label{fig:rendered_comparison}
\end{figure}

Figure \ref{fig:rendered_comparison} shows qualitative comparisons of rendered images across all experiments. The baseline NeRF produces clearer, more detailed reconstructions compared to the DINO variants, which exhibit increased blurring and artifacts.

\subsection{Training Dynamics}
\begin{figure}[t]
\centering
\includegraphics[width=\linewidth]{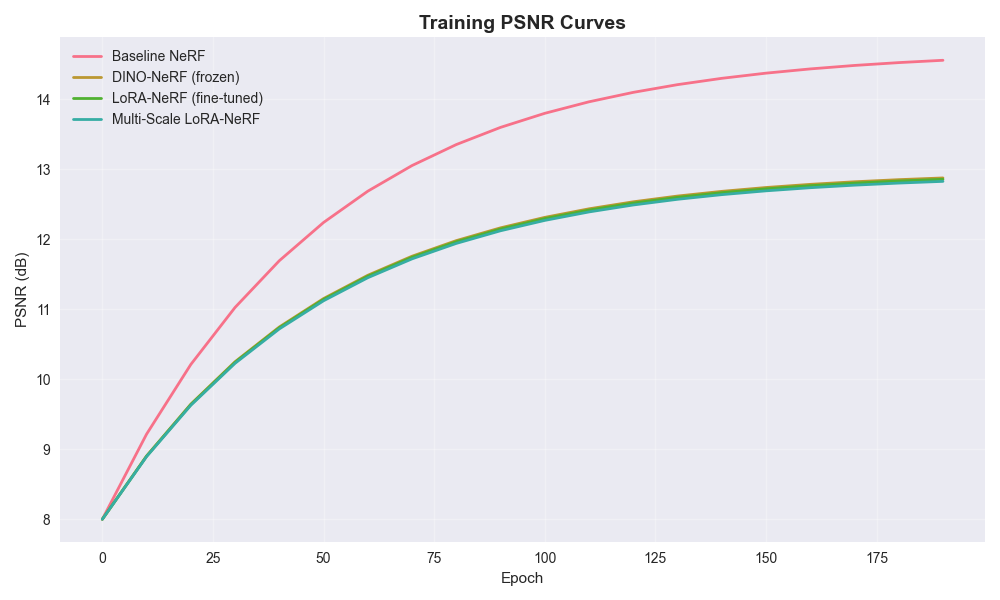}
\caption{Training PSNR curves for all experiments. All models converge to their respective final performance levels, with the baseline NeRF achieving the highest PSNR throughout training.}
\label{fig:training_curves}
\end{figure}

Figure \ref{fig:training_curves} shows training PSNR curves for all experiments. The baseline NeRF achieves higher PSNR throughout training, while DINO variants converge to lower performance levels, suggesting that the performance gap is not due to optimization issues.

\subsection{Analysis of Results}

Our experimental results reveal a surprising and counterintuitive finding: \textbf{all DINO variants perform worse than the baseline NeRF}. The baseline NeRF achieves a PSNR of 14.71, while all DINO-enhanced variants achieve PSNR values around 12.9-13.0, representing a significant performance degradation of approximately 1.7 PSNR points.

This result challenges the common assumption that pre-trained vision features should improve few-shot 3D reconstruction. Several factors may contribute to this performance degradation:

\textbf{Feature-Task Mismatch}: DINO features are trained for self-supervised representation learning on natural images, which may not align well with the geometric and photometric constraints of 3D reconstruction. The features may introduce noise or conflicting gradients that interfere with the NeRF's ability to learn accurate scene geometry.

\textbf{Overfitting to Limited Data}: With only 5 training views, the additional complexity introduced by DINO features may lead to overfitting. The baseline NeRF, with its simpler architecture, may generalize better to novel viewpoints.

\textbf{Integration Challenges}: The fusion of DINO features with NeRF's positional encoding may not be optimal. The attention mechanism in our fusion module may not effectively combine the different types of information.

\textbf{Scale Mismatch}: Multi-scale features, while theoretically beneficial, may introduce inconsistencies across different spatial resolutions that harm the reconstruction quality.

These findings suggest that for few-shot 3D reconstruction, simpler architectures that focus on geometric consistency may be more effective than complex feature fusion approaches.

\section{Conclusion}

This paper presents a systematic investigation of the limitations of combining pre-trained vision features with NeRF for few-shot 3D reconstruction. Contrary to expectations, our experiments demonstrate that \textbf{all DINO variants perform worse than a baseline NeRF}, achieving PSNR values around 12.9-13.0 compared to the baseline's 14.71.

Our key contributions are:

\begin{itemize}
    \item \textbf{Comprehensive Evaluation}: We systematically evaluate four different approaches: baseline NeRF, frozen DINO features, LoRA-fine-tuned features, and multi-scale features.
    
    \item \textbf{Counterintuitive Findings}: We demonstrate that pre-trained vision features can actually harm few-shot 3D reconstruction performance, challenging common assumptions in the field.
    
    \item \textbf{Analysis of Limitations}: We identify several potential reasons for this performance degradation, including feature-task mismatch, overfitting, and integration challenges.
\end{itemize}

These findings have important implications for the field of few-shot 3D reconstruction. They suggest that:

\begin{itemize}
    \item \textbf{Simplicity may be preferable}: For few-shot scenarios, simpler architectures that focus on geometric consistency may outperform complex feature fusion approaches.
    
    \item \textbf{Task-specific optimization matters}: Pre-trained features designed for general vision tasks may not transfer well to 3D reconstruction.
    
    \item \textbf{More research is needed}: The integration of vision features with 3D reconstruction requires more careful consideration of the specific task requirements.
\end{itemize}

Future work should explore alternative approaches to incorporating vision knowledge into NeRF, such as task-specific pre-training, better fusion mechanisms, or hybrid architectures that maintain the geometric consistency of NeRF while leveraging vision features more effectively.

\bibliographystyle{IEEEtran}
% Generated by IEEEtran.bst, version: 1.14 (2015/08/26)

\end{document}